\documentclass[runningheads,a4paper]{llncs}
\usepackage{amssymb}
\usepackage{graphicx}
\usepackage{times}
\usepackage{helvet}
\usepackage{courier}
\usepackage{algorithm}
\usepackage{algorithmicx}
\usepackage{algpseudocode}
\usepackage{amsmath}
\usepackage{latexsym}
\usepackage{booktabs}
\usepackage{caption}
\usepackage{url}
\graphicspath{{./pdf/}}
\DeclareGraphicsExtensions{.pdf}

\usepackage{url}
\urldef{\mailsa}\path|{lgy147, junlong, beethove, udars}@mail.ustc.edu.cn|
\urldef{\mailsb}\path|{ketang}@ustc.edu.cn|
\urldef{\mailsc}\path|{lgy147}@mail.ustc.edu.cn|
\urldef{\mailsd}\path|{minlonglin}@tencent.com|
\newcommand{\keywords}[1]{\par\addvspace\baselineskip
\noindent\keywordname\enspace\ignorespaces#1}

\begin{document}
\mainmatter  % start of an individual contribution
\title{Relief R-CNN : Utilizing Convolutional Features for Fast Object Detection}
\titlerunning{Relief R-CNN : Utilizing Convolutional Features for Fast Object Detection}
\author{Guiying Li\inst{1}%
%\thanks{The code of the framework can be found in https://github.com/IdiosyncraticDragon/relief\_rcnn. This work is partially supported by the USTCSCC.}%
\and Junlong Liu\inst{1}\and Chunhui Jiang\inst{1}\and Liangpeng Zhang\inst{1}\and Minlong Lin\inst{2}\and Ke Tang\inst{1}}
\authorrunning{G. Li et al.}
\institute{School of Computer Science and Technology,\\
University of Science and Technoloy of China,\\ Hefei, Anhui 230027, P.R. China\\
\mailsa\\
\mailsb\\
\and Tencent Company,\\Shenzhen, 518057, P.R. China\\
\mailsd\\
%\url{http://ubri.ustc.edu.cn}
}

\toctitle{Relief R-CNN}
\tocauthor{Guiying Li}
\maketitle

\begin{abstract}
R-CNN style methods are sorts of the state-of-the-art object detection methods, which consist of region proposal generation and deep CNN classification. However, the proposal generation phase in this paradigm is usually time consuming, which would slow down the whole detection time in testing. This paper suggests that the value discrepancies among features in deep convolutional feature maps contain plenty of useful spatial information, and proposes a simple approach to extract the information for fast region proposal generation in testing. The proposed method, namely Relief R-CNN ($R^2$-CNN), adopts a novel region proposal generator in a trained R-CNN style model. The new generator directly generates proposals from convolutional features by some simple rules, thus resulting in a much faster proposal generation speed and a lower demand of computation resources. Empirical studies show that $R^2$-CNN could achieve the fastest detection speed with comparable accuracy among all the compared algorithms in testing.
\end{abstract}

\section{Introduction}
\label{intro_section}
One type of the state-of-the-art deep learning methods for object detection is R-CNN \cite{girshick2014rich} and its derivative models \cite{girshickICCV15fastrcnn,ren2015faster}. R-CNN consists of two main stages: the category-independent region proposals generation and the proposal classification. The region proposals generation produces the rectangular Regions of Interest (RoIs) \cite{girshickICCV15fastrcnn,ren2015faster} that may contain object candidates. In the proposal classification stage, the generated RoIs are fed into a deep CNN \cite{krizhevsky2012imagenet}, which will classify these RoIs as different categories or the background.

However, R-CNN is time inefficient in testing, especially when running on hardwares with limited computing power like mobile phones. The time cost of R-CNN comes from three parts: 1) the iterative RoIs generation process \cite{Hosang2015Pami}; 2) the deep CNN with a huge computation requirement \cite{krizhevsky2012imagenet,simonyan2014very,he2015deep}; and 3) the naive combination of RoIs and the deep CNN \cite{girshick2014rich}. Many attempts on these three parts have been made to speed up R-CNN in testing. For RoI generation, Faster R-CNN \cite{ren2015faster} trains a Region Proposal Network (RPN) to predict RoIs in images instead of traditional data-independent methods that iteratively generate RoIs from images like Objectness \cite{alexe2010object}, Selective Search \cite{uijlings2013selective}, EdgeBox \cite{dollar2015fast} and Bing \cite{cheng2014bing}. For the time consuming deep CNN, some practical approaches \cite{han2015deep,kim2015compression} have been proposed to simplify the CNN structure. For the combination of RoIs and the deep CNN, SPPnet \cite{kaiming14ECCV} and Fast R-CNN \cite{girshickICCV15fastrcnn}, which are the most popular approaches, reconstruct the combination of RoIs and CNN by directly mapping the RoIs to a specific pooling layer inside the deep CNN model. However, all these methods still cannot be efficiently deployed on low-end hardwares, since they still require considerable computing.

In this paper, we propose Relief R-CNN ($R^2$-CNN), which aims to speed up the deployment of RoI generation for a trained R-CNN without any extra training. For a trained R-CNN style model in deployment phase, $R^2$-CNN abandons the original RoIs generation process used in training, and directly extracts RoIs from the trained CNN. $R^2$-CNN is inspired by the analogy between relief sculptures in real life and feature maps in CNN. Visualization of convolutional layers\cite{zeiler2014visualizing,mahendran2015understanding} has shown that convolutional features with high values in a trained CNN directly map to the recognizable objects on input images. Therefore, $R^2$-CNN utilizes these convolutional features for region proposal generation. That is done by directly extracting the local region wrapping features with high values as RoIs. This approach is faster than many other methods, since a considerably large part of its computations are comparison operations instead of time consuming multiplication operations. Furthermore, $R^2$-CNN uses the convolutional features produced by CNN for RoI generation, while most of the methods need additional feature extraction from raw images for RoIs. In short, $R^2$-CNN could reduce much more computations in RoI generation phase compared with other methods discussed above.

The rest of the paper is organized as follows: Section 2 describes the details of Relief R-CNN. Section 3 presents the experimental results about $R^2$-CNN and relevant methods. Section 4 concludes the paper.

\section{Relief R-CNN}
\label{method_section}
\begin{figure}
	\centering
	\includegraphics[scale=0.35]{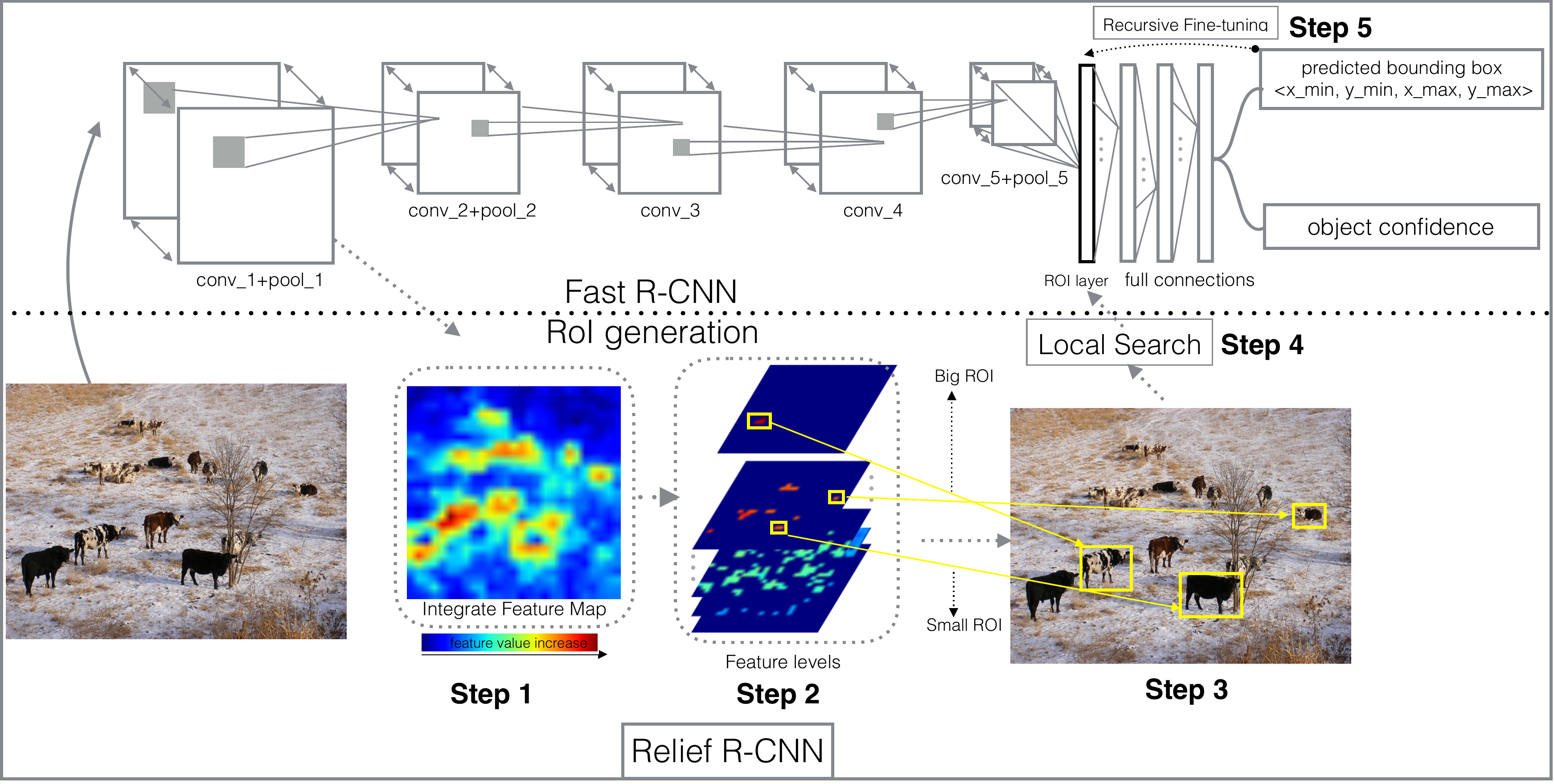}
	\caption{Overview of Relief R-CNN. (Step 1) First is generating an \textbf{Integrate Feature Map $f_\text{integrate}$} based on feature maps in pool1 layer of Alexnet \cite{krizhevsky2012imagenet}, (Step 2) followed by separating features of $f_\text{integrate}$ into different \textbf{Feature Levels}. (Step 3) Then extracting \textbf{Big RoIs} and \textbf{Small RoIs} and using (Step 4,5) additional proposal refinement techniques for better performance. The process conducted by solid lines is the procedure of Fast R-CNN, while the process along with dotted lines is the special work flow of $R^2$-CNN}
	\label{f1}
\end{figure}

In this section we present the details of $R^2$-CNN. Figure \ref{f1} shows the brief structure of $R^2$-CNN.

\textbf{General Idea}
\label{method_basic}
The value discrepancies among features in a feature map of CNN are sorts of edge details. These details are similar to the textures on sculpture reliefs, which describe the vision by highlighting the height discrepancies of objects. Intuitively speaking, two nearby features that have significant value discrepancy may indicate they are on the boundary of objects, which is a type of edge details. There comes the basic assumption of $R^2$-CNN: region proposals can be generated from the object boundaries, which consist of enough edge details described by significant value discrepancies in CNN feature maps, with some simple rules based on the characteristics of convolutional feature maps.

The idea above comes from the observations on convolutional feature maps \cite{zeiler2014visualizing,mahendran2015understanding}, and the similarity between the feature maps and sculpture relief, so that the proposed method is called Relief R-CNN. In testing phase, by searching the regions have significant more salient features than nearby context features in convolutional feature maps of a trained CNN, $R^2$-CNN can locate the objects in the source image by utilizing these region. $R^2$-CNN can be summarized into 5 steps as follows, in which steps 1$\sim$4 replace the RoI generator in the original trained models and step 5 boosts the performance of the fast generated RoIs in classification phase.

\textbf{Step 1. Integrate Feature Map Generation}
A synthetic feature map called \textbf{Integrate Feature Map}, denoted as $f_\text{integrate}$, is generated by adding all feature maps up to one map. $f_\text{integrate}$ brings two advantages, the first is dramatically reducing the number of feature maps, the second is eliminating noisy maps. The generation of $f_\text{integrate}$ consists of two steps:
\begin{itemize}
	\item[1] Each feature map is normalized by dividing by its maximal feature value.
	\item[2] A $f_\text{integrate}$ is generated by adding all the normalized feature maps together in element-wise.
\end{itemize}

\textbf{Step 2. Separating Feature Levels by Feature Interrelationship}
Once the $f_\text{integrate}$ is ready, feature levels in $f_\text{integrate}$ should be formulated. As wrote in \textbf{General Idea}, $R^2$-CNN tries to locate objects by a special sort of edge details, which is depicted by feature value discrepancies. However, it is hard to define how large the discrepancy between two features indicates a part of a boundary. To overcome this obstacle, we propose to separate features into different feature levels, and features in different feature levels are considered to be discriminative. Therefore, the contours formated by nearby features in a feature level directly represent the boundaries.

In this paper, feature levels in a $f_\text{integrate}$ are generated by dividing the value range of all the features into several subranges. Each subrange is a specific level which covers a part of features in the $f_\text{integrate}$. The number of subranges is a hyper-parameter, denoted as $l$. $R^2$-CNN uniformly divides the $f_\text{integrate}$ into $l$ feature levels, see Algorithm \ref{pseudo_feature_level}. The step 2 in Figure \ref{f1} shows some samples of feature levels generated from the first pooling layer of CaffeNet model (CaffeNet is a caffe implementation of AlexNet \cite{krizhevsky2012imagenet}).

\renewcommand{\algorithmicensure}{\textbf{Input:}}
\begin{algorithm}
	\caption{Feature Level Separation}
	\label{pseudo_feature_level}
	\begin{algorithmic}[1]
		\Ensure ($f_\text{integrate}$, $l$) $\ \ \ \rhd$Integrate Feature Map and Feature Level Number
		\State Finding the maximal value $value_\text{max}$ and minimal value $value_\text{min}$ in $f_\text{integrate}$
		\State $\rhd$uniformly dividing the value range into $l$ subranges%$\rhd$Value range is $(value_{max} - value_{min})$, 
		\State $stride = (value_\text{max} - value_\text{min})/l$ %$\rhd$uniformly dividing the value range into $l$ subranges
		\State $\rhd feature_\text{level\_i}$ is the feature level $i$ for $f_\text{integrate}$%$\rhd\ \ \ $features in each subrange form a feature level
		\For {$i=1 \to l$}%$\ \ \rhd feature_{level\_i}$ is the feature level $i$ for $f_\text{integrate}$
		\State Finding features bigger than $value_\text{min}+(i-1)*stride$ and smaller than $value_\text{min}+i*stride$ in $f_\text{integrate}$ as $feautre_\text{level\_i}$
		%\State $feature_{level\_i}$ is the feature level $i$ for $f_\text{integrate}$
		\EndFor
		\State \Return {$<feature_\text{level\_1},...,feature_\text{level\_l}>$}
	\end{algorithmic}
\end{algorithm}

\textbf{Step 3. RoIs Generation}
The approach $R^2$-CNN adopted for RoIs generation is, as be mentioned in step 2, finding the contours formated by nearby features in a feature level, which needs the help of some deep network structure related observations. As the step 3 shown in Figure \ref{f1}, the neighboring features, which are surely belong to the same object, can form a small RoI. Furthermore, a larger RoI can be assembled from several small RoIs, in case of some large objects be consisted of small ones. Here's the summarized operations:
\begin{itemize}
	\item Small RoIs:
	Firstly, it searches for the feature clusters (namely the neighboring features) in the given $feature_\text{level\_i}$, and then mapping the feature clusters to the input image as \textbf{Small RoIs}.
	\item Big RoI:
	For the purpose of simplicity (avoiding the combinatorial explosion), only one \textbf{Big RoI} is generated in a feature level by assembling all the small RoIs.
\end{itemize}
%\begin{figure}
%	\centering
%	\includegraphics[scale=0.25]{riid_figure_3.pdf}
%	\caption{Overview of step 1$\sim$step 3. Step 1(left-bottom part): adding up all feature maps in a layer as $f_\text{integrate}$. Step 2(right-bottom part): Feature levels separating shows the value distribution of features in a feature map, the features with similar values in a feature map are grouped in the same feature level. Step 3(right-top part): There are many \textbf{Small RoIs} in each feature level since small objects may appear as feature clusters in the feature level. One \textbf{Big RoI} for each feature level can be taken as composition of all small RoIs. $f_\text{integrate}$ is uniformly separated into ten feature levels in experiment.}
%	\label{f3}
%\end{figure}

\textbf{Step 4. Local Search}
Convolutional features from source image are not produced by seamless sampling. As a result, RoIs extracted in convolutional feature maps might be quite coarse. Local Search in width and height is applied to tackle this problem. For each RoI, which its width and height are denoted as $(w, h)$, local search algorithm needs two scale ratios $\alpha$ and $\beta$ to generate 4 more RoIs: $(\beta*w, \beta*h), (\beta*w, \alpha*h), (\alpha*w, \alpha*h), (\alpha*w, \beta*h)$. In experiments, $\alpha$ was fixed to $0.8$ and $\beta$ was fixed to $1.5$. The Local Search can give about $1.8$ mAP improvement in detection performance.

\textbf{Step 5. Recursive Fine-tuning}
Previous steps provide a fast RoI generation for testing. However, the accuracy of testing is restricted because of the different proposals distribution between training and testing. Owing to this fact, we propose the method called recursive fine-tuning to boost the detection performance during the classification phase of RoIs.

The recursive fine-tuning is a very simple step. It does not need any changes to existing R-CNN style models, but just a recursive link from the output of a trained box regressor back to its input. Briefly speaking, it is a trained box regressor wrapped up into a closed-loop system from a R-CNN style model. This step aims at making full use of the box regressor, by recursively refining the RoIs until their performance have been converged. 

It should be noticed that there exists a similar method called Iterative Localization \cite{gidaris2015object}. It needs a bounding box regressor be trained in another settings and starts the refinement from the proposals generated by Selective Search, while the recursive fine-tuning bases on the regressor in a unified trained R-CNN and starts refinement from the RoIs generated by above steps (namely Step 1$\sim$4). Furthermore, recursive fine-tuning does not reject any proposals but only improve them if possible, while iterative localization drops the proposals below a threshold at the beginning.

\section{Experiments}
\label{experiments_section}
\subsection{Setup}
In this section, we compared our $R^2$-CNN with some state-of-the-art methods for accelerating trained R-CNN style models.
The proposals of Bing, Objectness, EdgeBoxes and Selective Search were the pre-generated proposals published by \cite{Hosang2015Pami}, since the the algorithm settings were the same. The evaluation code used for generating Figure \ref{f6} was also published by \cite{Hosang2015Pami}.

The baseline of R-CNN style model is Fast R-CNN with CaffeNet. The Fast R-CNN model was trained with Selective Search just the same as in \cite{girshickICCV15fastrcnn}. The Faster R-CNN \cite{ren2015faster} used in experiments was based on project py-faster-rcnn \cite{py_faster_rcnn_project}.
 Despite the difficulty of Faster R-CNN for low power devices, RPN of Faster R-CNN is still one of the state-of-the-art proposal methods. Therefore, RPN was still adopted in experiments using the same Fast R-CNN model consistent with other methods for detection. The RPN in experiments was trained on the first stage of Faster R-CNN training phases. This paradigm is the unshared Faster R-CNN model mentioned in \cite{ren2015faster}.
 For the $R^2$-CNN model, the number of recursive loops was set as 3, and the number of feature levels was 10. 

All experiments were tested on PASCAL VOC 2007 \cite{everingham2014pascal}. Deep CNNs in this section got support from Caffe \cite{jia2014caffe}, a famous open source deep learning framework.
 All the proposal generation methods were running on CPU (inc. $R^2$-CNN and RPN) while the deep neural networks of classification were running on GPU. All the deep neural networks had run on one NVIDIA GTX Titan X, and the CPU used in the experiments was Intel E5-2650V2 with 8 cores, 2.6Ghz.

\subsection{Speed and Detection Performance}
\begin{table}[tp]
	\scriptsize
	\caption{Testing Time \& Performance comparison. The object detection model used here is Fast R-CNN. The $R^2$-CNN needs recursive fine-tuning which makes classification be time-consuming. ``Total Time" is the sum of values in ``Proposal Time" and ``Classification Time" . ``*" indicates the runtime reported in \cite{Hosang2015Pami}. ``RPN" is the proposal generation model used in Faster R-CNN. \textbf{Bold} items are the results of $R^2$-CNN. $R^2$-CNN presents the fastest speed and comparable detection performance.}
	\label{t6}
	\centering
	\begin{tabular}{l|cccccc}
		\toprule
		Methods & Proposal Time (sec.) & Proposals &Classification Time (sec.)& Total Time (sec.) &  mAP & mean Precision (\%)\\
		\midrule
		$R^2$-CNN & \textbf{0.00048} & \textbf{760.19} & \textbf{0.146} & \textbf{0.14648} & \textbf{53.8} & \textbf{9.2}\\
		Bing & 0.2* & 2000 & 0.115 & 0.315 &41.2 & 2\\
		EdgeBoxes & 0.3* & 2000 & 0.115 & 0.415& 55.5 & 4.2\\
		%RPN  & 1.445 & 2000 & 0.115 & 1.560& 50.9 & 8.2\\
		RPN  & 1.616 & 2000 & 0.115 & 1.731& 55.2 & 3.5\\%with nms
		Objectness & 3* & 2000 & 0.115 & 3.115&  44.4 & 1.7\\
		Selective Search & 10* & 2000 & 0.115 & 10.115& 57.0 & 5.9\\
		%0.2757
		\bottomrule
	\end{tabular}
\end{table}
Table \ref{t6} contains the results of comparison about time in testing. The testing time is separated into proposal time and classification time. The proposal time is the time cost for proposal generation, and the classification time is the time cost for verifying all the proposals.

%\subsection{Detection Performance}
Table \ref{t6} has also shown the detection performances of $R^2$-CNN and other comparison methods. Precision \cite{ozdemir2010performance} is a well known metric to evaluate the precision of predictions, mAP (abbreviation of mean Average Precision) is a highly accepted evaluation in the object detection task \cite{russakovsky2014imagenet}.

The empirical results in Table \ref{t6} reveal that $R^2$-CNN could achieve a very competitive detection performance compared with state-of-the-art Selective Search, EdgeBoxes and Faster R-CNN with a much more fast CPU speed, which means it's a more suitable RoI method for deploying trained R-CNN style models on low-end hardwares.

\subsection{Proposal quality}
\label{riid_compare_to_other_mehtods}
\begin{figure}
	\centering
	\includegraphics[scale=0.9]{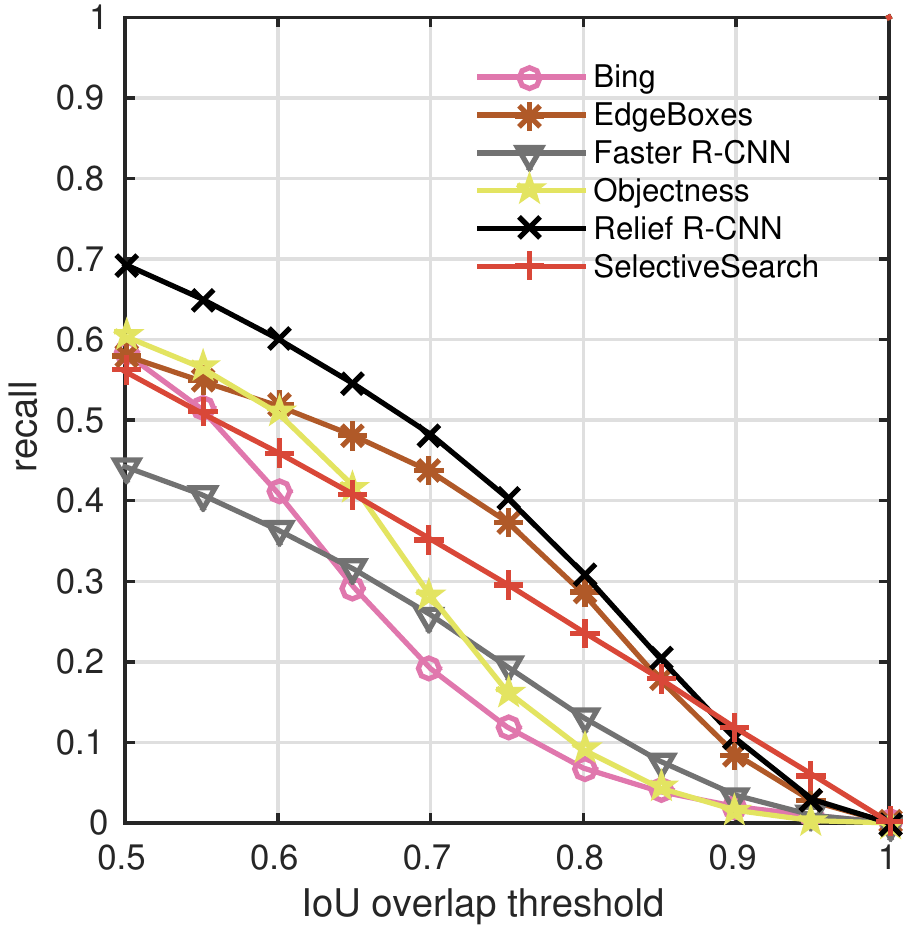}
	\caption{Recall to $IoU$ threshold with 200 proposals in count. $R^2$-CNN had nearly dominated other methods. }
	\label{f6}
\end{figure}

To evaluate the quality of proposals, the evaluation metric \cite{Hosang2015Pami} Recall-to-IoU curve was adopted, see Figure \ref{f6}. The metric $IoU$ (abbreviation of $intersection\ over\ union$) \cite{russakovsky2014imagenet}, is an evaluation criterion to measure how similar two regions are. A larger $IoU$ indicates more similar regions.

In Figure \ref{f6}, it could be found that $R^2$-CNN had nearly dominated other methods in $IoU$ threshold between 0.5$\sim$0.9, and became the secondary best in $IoU$ threshold 0.9$\sim$1.0.

It should be noticed that $R^2$-CNN could not control the number of proposals, but it got the best results with hundreds of proposals while others need thousands. The experiments in this section have shown that $R^2$-CNN could get a very good performance in the situation of limit proposals with a high speed, which is also a good character for platforms with limited computation resources.

\section{Conclusion}
\label{conclusion_and_future_section}
This paper presents a unified object detection model called Relief R-CNN ($R^2$-CNN). By directly extracting region proposals from convolutional feature discrepancies, namely the location information of salient features in local regions, $R^2$-CNN reduces the RoI generation time required for a trained R-CNN style model in testing phase. Hence, $R^2$-CNN is more suitable to be deployed on low-end hardwares than existing R-CNN variants. Moreover, $R^2$-CNN introduces no additional training budget. Empirical studies demonstrated that $R^2$-CNN was faster than previous works with competitive detection performance. 

\subsubsection*{Acknowledgments.}
This work was supported in part by the National Natural Science Foundation of China under Grant 61329302 and Grant 61672478, and in part by the Royal Society Newton Advanced Fellowship under Grant NA150123.

%\bibliographystyle{splncs03}
%\bibliography{RIID}

\end{document}